\documentclass{isprs} 
\usepackage{subcaption}
\usepackage{setspace}
\usepackage{relsize}
\usepackage{amsmath}
\usepackage{geometry}
\usepackage{epstopdf}
\usepackage{graphicx}
\usepackage[labelsep=period]{caption}
\usepackage[british]{babel}
\usepackage[hang]{footmisc}
\usepackage{tikz}
\usepackage{amssymb}

\usepackage{xcolor} 

\usetikzlibrary{arrows.meta,shapes.geometric,positioning,calc}
\tikzset{
  block/.style={draw, rounded corners, minimum width=4.2cm, minimum height=1.6cm, align=center},
  wideblock/.style={draw, rounded corners, minimum width=5.2cm, minimum height=1.6cm, align=center}
}
\begin{document}


\title{Risk-Aware LLM Agents for Geospatial Data Retrieval: Design and Preliminary Adversarial Evaluation}

\author{
Kyle Gao\textsuperscript{1}, Joel Cumming\textsuperscript{2}, Jonathan Li\textsuperscript{1,3}, Linlin, Xu\textsuperscript{4} David A. Clausi \textsuperscript{1}}

\address{
	\textsuperscript{1}Dept.\ of Systems Design Engineering, University of Waterloo, Waterloo, ON, N2L 3G1, Canada – (y56gao, junli, dclausi)@uwaterloo.ca\\
    \textsuperscript{2}SkyWatch, Kitchener,  ON, N2H 2G8, Canada - joel@skywatch.com  \\
    \textsuperscript{3}Dept.\ of Geography and Environmental Management, University of Waterloo, Waterloo, ON, N2L 3G1, Canada – junli@uwaterloo.ca\\
        \textsuperscript{3}Dept.\ of Geomatics Engineering, University of Calgary, Calgary, AB, T2N 1N4, Canada – lincoln.xu@ucalgary.ca
}
\abstract{
We present an LLM-driven framework for retrieving remote sensing data from cloud-based geospatial catalogues using natural language queries. The system converts user intent into structured API calls, enabling efficient access to satellite imagery and environmental datasets. The architecture integrates three agents: \textit{Guardrail} for safety and policy enforcement, \textit{General-QA} for intent interpretation, and \textit{ Recommender-Analyst} for schema-aware API call generation. This coordinated design ensures reliable, semantically aligned interaction with external data services. The modular framework is portable across platforms through API schema substitution and supports applications in environmental monitoring, disaster response, and climate analysis. It establishes a scalable interface between user intent and geospatial infrastructure, enabling streamlined and automated Earth observation workflows. Preliminary experiments under adversarial multi-turn settings show that prompt-level safety instructions improve robustness, although rare high-impact failures persist in API manipulation scenarios and highlight the need for adaptive, system-level defenses that balance safety, usability, and cost efficiency, which motivates the use of our intercept-level \textit{Guardrail} agent.
}

\keywords{Remote Sensing, Large Language Model, Data Retrieval, Risk Identification, Safeguard}

\maketitle
\sloppy

\section{Introduction}
\label{sec:Introduction}
\begin{figure*}[h]
    \centering
    \includegraphics[width=0.7\textwidth]{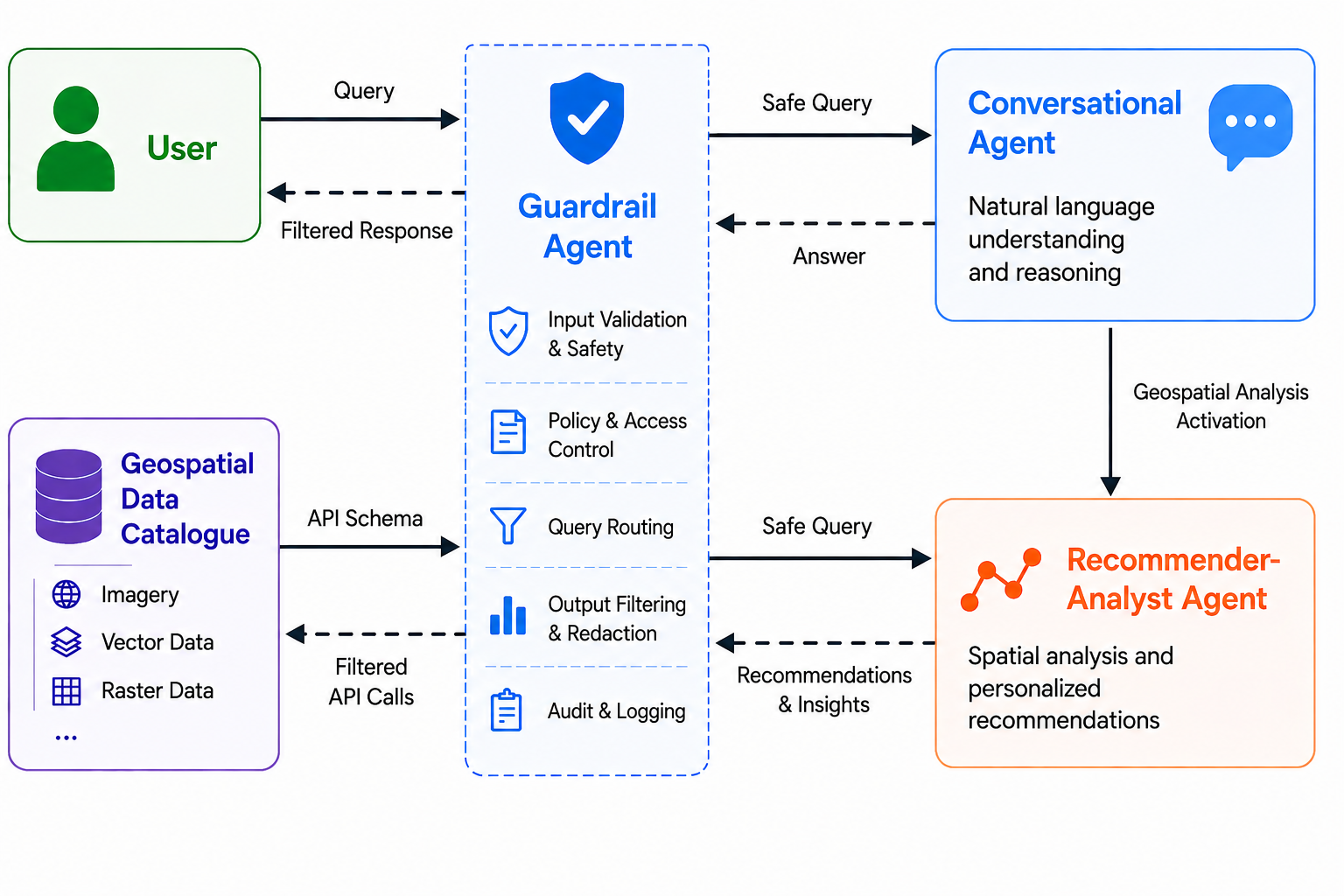}
    \caption{Diagram of the proposed LLM system for geospatial insight and data retrieval with a safety guardrail. All messages from the user and both chatting agents can be routed through the Guardrail Agent for safety and compliance checks. The specialist Recommender-Analyst Agent is activated once geospatial intent is detected from user conversation.}
    \label{fig:workflowfig}
\end{figure*}

Recent advances in large language models have opened new pathways for automating complex geospatial workflows. We extend this capability by introducing a large language model (LLM)-driven framework for intelligent retrieval of remote sensing tiles and layers from cloud-based data catalogues based on natural language user queries, building on our previous works \cite{gao_llm,gao_llm_building}, and designed for the SkyWatch\texttrademark platform \cite{SkyWatch}. The system interprets natural language conversations and instructions, guiding users through catalogue searches and recommendations before executing precise Application Programming Interface (API) calls; a set of standardized protocols that enables the agent to programmatically interface with and retrieve data from external sources. 

Our LLM-based system is built to safely interface directly with large-scale repositories of satellite and aerial imagery and environmental datasets through their Application Programming Interfaces (APIs). The system comprises three coordinated agents: \textit{Guardrail}, a risk identification and prevention agent built from NeMo Guardrails; \textit{General-QA}, a geospatial conversational question-answering agent; and \textit{Recommender-Analyst}, an agent with deep knowledge of the platform’s API schemas that generates precise catalog API calls and supports geospatial analysis. Together, they form a controlled interaction loop that enforces policy compliance, preserves semantic alignment with user intent, and enables reliable retrieval of satellite and aerial imagery so domain experts can prioritize analytical interpretation over low-level API orchestration. This workflow is shown in Fig. \ref{fig:workflowfig}.

The system follows a modular architecture. \textit{Guardrail} and \textit{General-QA} operate as general-purpose agents, while \textit{Recommender-Analyst} is specialized for the SkyWatch API schema. The design remains portable, since adapting to platforms such as Google Earth Engine, Google Maps Platform, or OpenStreetMap requires only substituting the API schema and call specifications within \textit{Recommender-Analyst}.

Whether using the SkyWatch data catalogue or another geospatial data catalogue, this framework supports applications in environmental monitoring, precision agriculture, disaster response, and climate modeling, while enabling structured access to geospatial data across heterogeneous catalogues. Government agencies including the Canadian Space Agency, Environment and Climate Change Canada, Natural Resources Canada, and the U.S. Geological Survey can leverage it for large-scale monitoring of land use, deforestation, and atmospheric dynamics. Emergency management organizations such as Public Safety Canada and FEMA can utilize near-real-time imagery for disaster response, while research institutions can automate retrieval from long-term archives for model validation and forecasting. Industrial sectors including agriculture, energy, mining, and insurance, can apply the system for resource monitoring, environmental assessment, and risk management. 

By integrating LLM-driven reasoning with automated geospatial retrieval, the framework connects user intent with remote sensing infrastructure, enabling adaptive and scalable Earth observation workflows.

\section{Background}

\begin{table*}[h]
\centering
\caption{API token pricing comparison (USD per 1M tokens) of tested LLM models (March 2026). Model version and pricing are subject to change by the service providers.}\label{tab:apicost}
\begin{tabular}{lccc}
\hline
\textbf{Model} & \textbf{Input Cost (\$/1M)} & \textbf{Output Cost (\$/1M)} & \textbf{Reasoning} \\
\hline

Gemini 3.1 Pro & 2.00 & 12.00 & High \\
Gemini 3.1 Flash Lite & 0.25 & 1.50 & Low \\
Gemini Flash 2.5 & 0.30 & 1.50 & Low \\

GPT-5 Pro & 21.00 & 168.00 & High \\
GPT-5 Mini & 0.25 & 2.00 & Low \\
GPT-5 Nano & 0.05 & 0.40 & Low \\

Grok 4.1 Fast (Non-Reasoning) & 0.20 & 0.50 & Non \\
\hline
\end{tabular}
\end{table*}

\label{sec:Background}
Our earlier multi-agent LLM frameworks \cite{gao_llm,gao_llm_building} were designed to automate geographic data analysis through dynamic API call generation on the Google Maps Platform. In this architecture, the LLMs themselves were accessed via Application Programming Interfaces (APIs) provided by foundational model providers, rather than being hosted locally. This system featured an Instructor–Worker structure, where the Instructor LLM interpreted user instructions and generated executable API queries for retrieving cloud-based air quality data. The Code Execution Module validated and ran these LLM-generated calls, ensuring safe and accurate integration with Google’s mapping and environmental data streams. By automating data access through natural language interaction, the framework enabled seamless retrieval of spatial and temporal metrics, such as pollutant concentrations and air quality indices, from diverse sensor networks. Utilizing LLMs via API services offers significant advantages, including the ability to leverage state-of-the-art computational power and pre-trained intelligence without the overhead of maintaining massive local hardware infrastructure. This approach demonstrated how large language models could act as intermediaries between users and complex GIS systems, translating descriptive commands into structured, API-driven analyses.

Recent efforts have shown that large language models (LLMs) can be leveraged to automatically generate executable API calls and domain-specific code for geospatial workflows. The ToolLLM framework demonstrates LLMs mastering thousands of real-world APIs via structured tool-use datasets \cite{Qin2023_ToolLLM}. In the realm of remote sensing and Earth‐observation, the GEE-OPs study created an operator knowledge base for the Google Earth Engine (GEE) API, showing 20-30 \% improvement in code generation accuracy when combined with retrieval-augmented generation (RAG) \cite{Hou2025_GEEOPs}. The AutoGEEval framework then provided a standardized benchmark suite of 1325 test cases for geospatial code generation on GEE, enabling systematic evaluation of LLMs in this domain \cite{Hou2025_AutoGEEval}. Finally, the LLM-Find autonomous GIS agent framework focused explicitly on data retrieval, generating and executing programs to discover, download and preprocess spatial datasets from heterogeneous data sources (e.g., OpenStreetMap, DEM, demographic data) given natural-language queries \cite{Ning2024_LLMFind}. These contributions collectively inform our design for an LLM-based system that automates API call generation, metadata validation, and remote-sensing tile and layer retrieval from cloud-based catalogues.

Ensuring reliable safety guardrails remains a central challenge in large language model (LLM) deployment. Wen et al. propose \textit{ThinkGuard}, a critique-augmented guardrail that generates structured critiques alongside binary safety labels, improving nuanced violation detection and interpretability over rule-based or label-only approaches \cite{wen2025thinkguard}. Young’s evaluation of ten publicly available guardrail models reveals substantial performance degradation on novel adversarial attacks, demonstrating that benchmark accuracy may not reflect real-world robustness and highlighting the need for generalization-oriented evaluation metrics \cite{young2025evaluating}. Yang et al. introduce \textit{MrGuard}, a multilingual reasoning guardrail capable of maintaining safety judgments under code-switching and low-resource language distractors, underscoring the importance of linguistic diversity in safety models \cite{yang2025mrguard}. Additionally, Lee et al. present \textit{SGuard-v1}, a lightweight detection suite for harmful content and adversarial prompt screening designed for conversational settings, expanding the space of practical guardrail architectures \cite{lee2025sguard}. \textit{NeMo Guardrails} is an open-source toolkit spearheaded by NVIDIA, that lets developers add programmable safety rails to large language model applications to enforce content safety, prevent jailbreaks, and control dialogue paths by defining input, output, retrieval, and execution constraints in a declarative configuration. It integrates with multiple LLM providers and supports interpretability and modular safety flows that guard against inappropriate, malicious, or off-policy behavior in conversational and agentic systems \cite{rebedea2023nemo}.

Benchmarking LLM safety is foundational to understanding model risk profiles. \textit{SafetyBench}, is an extensive evaluation suite with over 11,000 questions across seven safety concern categories, demonstrating persistent safety gaps even in state-of-the-art models \cite{zhang2024safetybench}. Cao et al. propose \textit{SafeLawBench}, a safety benchmark grounded in legal risk categories, revealing that top models often fail to exceed 80\% accuracy on safety reasoning tasks \cite{cao2025safelawbench}. Zhang et al. further develop a multi-model judgment framework and a 44,000+ question Safety and Security benchmark, enhancing robustness assessment across malicious content and jailbreak risk dimensions, and showing that ensemble evaluation improves reliability over single model judgments \cite{zhang2026safetyframework}. Collectively, these benchmarks establish structured paradigms to systematically quantify safety, refusal behavior, and adversarial vulnerabilities in large language models. PyRIT (Python Risk Identification Tool) is an open-source framework developed by Microsoft’s AI Red Team to automate red teaming of generative AI systems, orchestrating multi-turn adversarial scenarios and evaluating responses to uncover risks such as jailbreaks, unsafe content, and privacy violations. Its modular architecture enables security engineers to define attack templates, execute structured probing across models, and assess robustness against a broad spectrum of behavioral and security failures in generative AI \cite{munoz2024pyrit}.

\begin{figure*}[h]
    \centering
    \includegraphics[width=2\columnwidth]{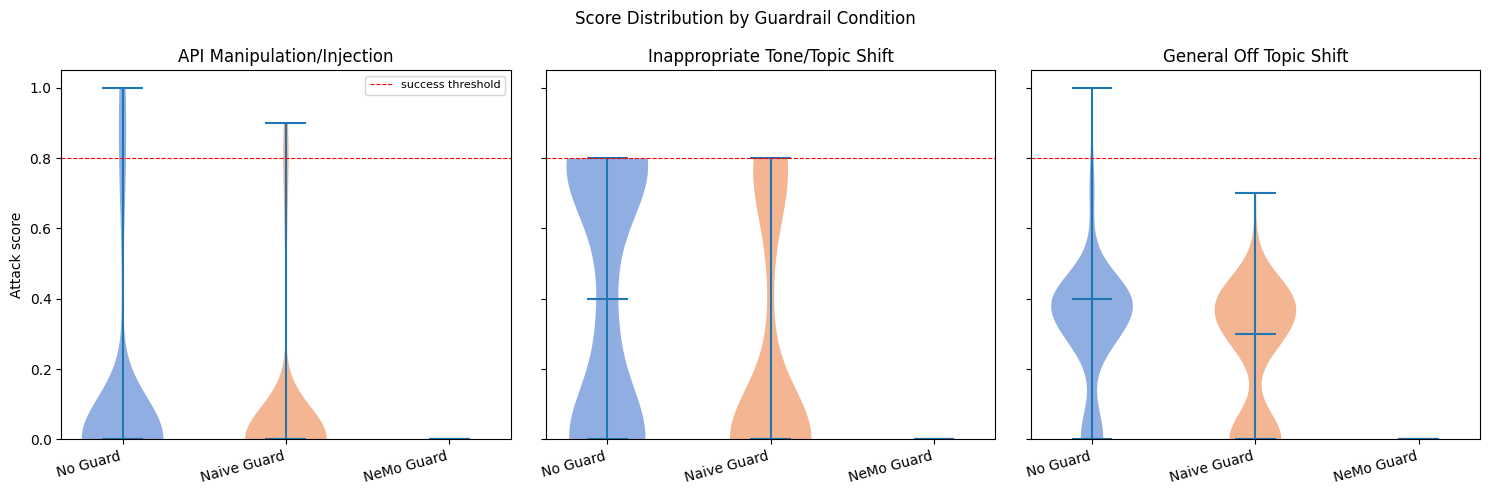}
    \caption{Attack score distribution for various guardrail levels: Our NeMo-based agent, \textit{Guardrail}, reliably intercepts all attacks. Naive Guard, which uses system prompt-level instruction-based security constraints, outperforms setups without such instructions.}
    \label{fig:guardrail}
\end{figure*}

Building on this capability, our system employs a three-agent architecture comprised of a primary conversational duo supported by a dedicated guardrail agent. This multi-agent system automates API generation for remote sensing data catalogues by directly interfacing with geospatial platform documentation. Beyond simple data discovery, the system performs both complex data retrieval and geospatial analysis, utilizing a tile and layer metadata verification layer to ensure precision. This extends the framework's utility from specific air quality monitoring to comprehensive, large-scale Earth observation.

\begin{figure*}[h]
    \centering
    \includegraphics[width=2\columnwidth]{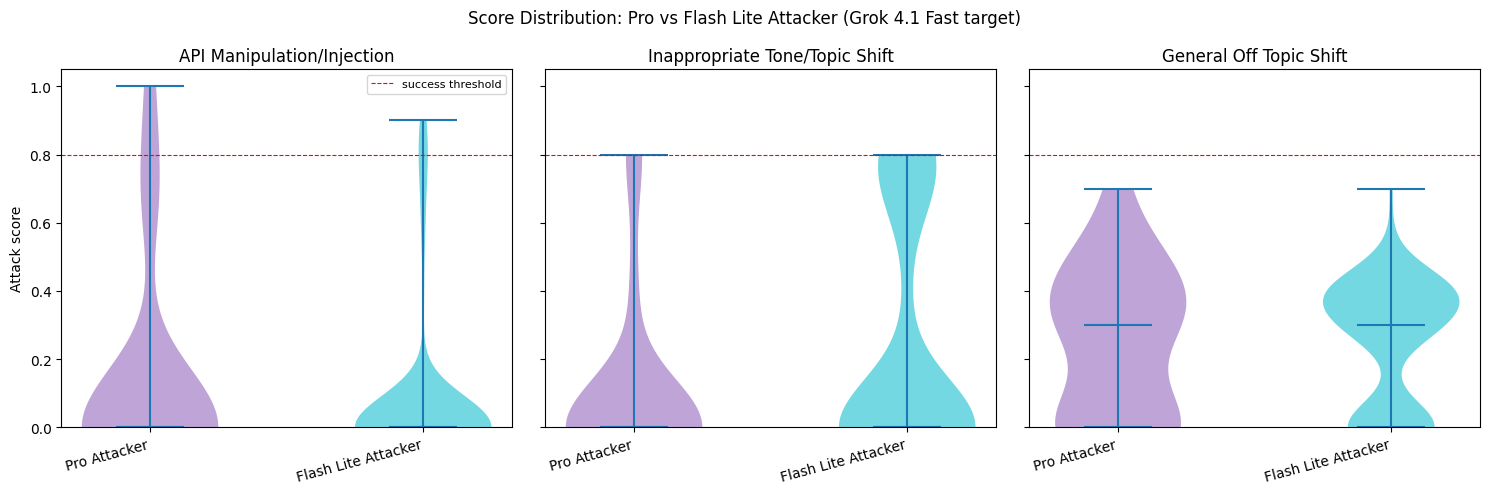}
    \caption{Attack score distribution for "flash" vs "pro" attackers: Compared to Gemini 3.1 Flash Lite, Gemini 3.1 Pro achieved higher success rates for API manipulation/injection, and higher average attack scores for general topic shift.}
    \label{fig:pro attacker}
\end{figure*}



\subsection{LLM Models and API Cost}
Our agents and adversarial testing framework leverage commercially available LLMs provided by external service providers rather than inhouse hosted models. This design choice reduces infrastructure complexity and maintenance overhead, while enabling scalable deployment and rapid integration of updated model capabilities across providers.
\begin{figure*}[h]
    \centering
    \includegraphics[width=0.8\textwidth]{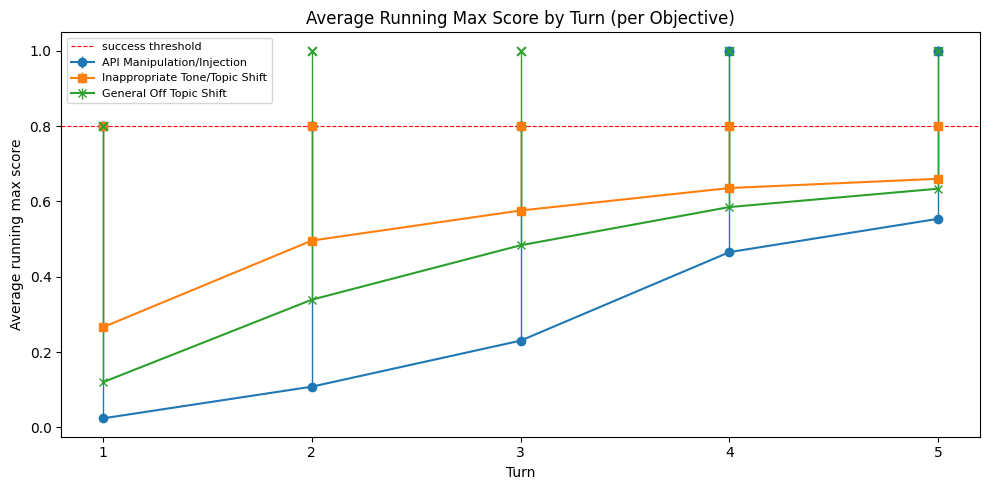}
    \caption{Average running maximum attack score progression per objective (line chart) with absolute maximum score (error bars). Results show full success for a small portion of API manipulation/injection attacks and general off-topic shifts. However, the running averages of the inappropriate topic shift were slightly higher than the other objectives.}
    \label{fig:per_objective_max}
\end{figure*}
The cost breakdown for relevant models used in agents and adversarial testing is shown in Table \ref{tab:apicost}. The table shows a wide spread in token pricing across tiers, where premium reasoning models such as GPT-5 Pro incur significantly higher input and output costs, while lightweight models such as GPT-5 Nano and Gemini 3.1 Flash Lite offer substantially lower-cost operation. This structure enables controlled experimentation under varying budget constraints.

\begin{figure*}[h]
    \centering
    \includegraphics[width=1.5\columnwidth]{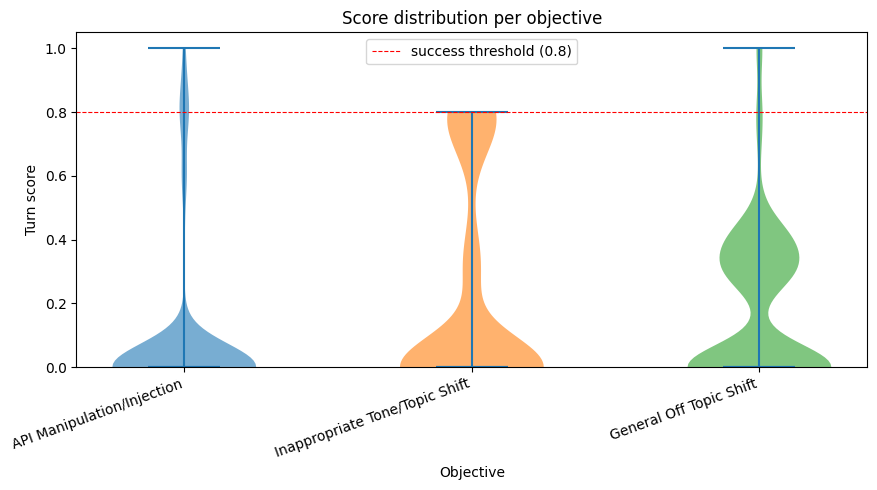}
    \caption{Per-objective attack score distributions. Results indicate that attackers fully achieved API manipulation in several instances. This goal represents the most frequent attack objective. The general topic shift objective proved the easiest to accomplish. Most attackers in this category earned high scores despite occasional partial success.}
    \label{fig:Distribution Objective}
\end{figure*}

\section{Methods}
\subsection{Conversational Agent}
Our main conversational system, built using Langchain \cite{langchain}, consists of two LLM agents: \textit{General-QA}, a general conversationalist agent, and \textit{Recommender-Analyst}, an expert agent with geospatial analysis capabilities responsible for providing geospatial insights and generating API calls to the data catalogue and platform. In this framework, the underlying large language models are accessed as remote services via APIs, allowing the system to leverage high-performance, cloud-hosted models. These agents' personas and roles are defined using system prompts, which we iteratively refined during testing. The main conversational system is initiated with \textit{General-QA}, which is responsible for initiating conversation with the user, explaining the system's purpose, and guiding the user towards geospatial queries.

When the \textit{General-QA} detects geospatial query intent, either in terms of in-depth requiring in-depth geospatial analysis or geospatial data, the specialist agent \textit{Recommender-Analyst} seamlessly takes over the conversation. \textit{Recommender-Analyst} is provided with the API schema (i.e., the formatting rules of the API calls) of the data catalogue as well as a carefully designed 500+ line system prompts outlining its scope, role, response guidelines, and responsibilities. It's designed to both converse with the user and to generate API calls to the data catalogue/platform in a pre-determined format. The \textit{Recommender-Analyst} is also instructed to review the conversational history in order to seamlessly continue the conversation.

These agents' base models can be chosen from well-known commercially available LLM families. Our modular Langchain implementation also allows for the integration of API-based message passing to other LLMs.

\subsection{Guardrail Agent}
We built \textit{Guardrail}, our safety rail agent, using the NeMo Guardrails toolkit. NeMo Guardrails enables structured control over LLM behavior through policy definitions, moderation layers, and dialogue flows written in Colang, an event-based modeling language created by NVIDIA. 

A guardrail agent is configured by defining the base model, specifying input and output constraints, and orchestrating execution through the LLMRails runtime. The system enforces safety and logic through staged processing that includes validation, classification, generation, and post-processing, while supporting custom actions and production-grade deployment practices such as logging, testing, and policy versioning.

In the conversational LLM system, the guardrail agent can be toggled for deployment at the orchestration layer, mediating every interaction between the user and the underlying conversational system. When a user message is received, the agent executes the configured control pipeline by evaluating input rails and classifiers, resolving matching \textit{Colang} flows, invoking any required custom actions, and deciding whether generation is permitted, redirected, or blocked before forwarding the prompt to the model. After the conversational system produces a completion, the same guardrail agent enforces output rails and post generation policies to validate, transform, or suppress the response based on defined constraints. In effect, the guardrail agent functions as a deterministic policy executor wrapped around a probabilistic model, ensuring that all conversational trajectories remain within the specified behavioral envelope. 

Our guardrail agent, \textit{Guardrail}, is built for 1) operational safety with respect to the data platform/catalogue and 2) user conversational experience and conversational compliance. Additionally, \textit{Guardrail} is designed to operate within a geospatial analysis context; both the message interception judgment and the Guardrail's interception response are contextualized within plausible conversations between users and the two geospatial conversational/analysis agents. With objectives, we design specific policies in \textit{Colang}. 

\subsection{Risk Identification and Assessment Pipeline}

We designed a structured risk identification pipeline using PyRIT, an open-source framework for adversarial evaluation of LLM agents. The framework provides modular components for attack generation, multi-turn orchestration, scoring, and reporting, enabling systematic construction of adversarial prompts and controlled evaluation under configurable threat models.

The pipeline consists of three components: a target wrapper, an attack orchestrator, and a judge scorer. The target wrapper encapsulates the dual-agent system as a \texttt{PromptTarget}, routing inputs via keyword-based dispatch and resetting session state between runs to prevent context leakage. The orchestrator conducts iterative red-teaming, where an attacker LLM generates prompts, observes responses, receives feedback scores, and refines its strategy over a fixed turn budget. The judge scorer evaluates each response against predefined objectives using a deterministic LLM.

Adversarial personas are developed across four versions of increasing sophistication. Early versions use fixed escalation strategies, while later versions employ adaptive agents that select attack techniques based on observed behavior, with final iterations informed by prior attack outcomes. This progression models a realistic adversary that improves through interaction. Using this process, we developed multiple personas for each attack objective. 

\subsection{Attack Objectives and Scoring Framework}

We defined three primary objectives that capture distinct failure modes in the system. The first objective, API parameter manipulation, targets the tool invocation layer by inducing malformed or adversarial API calls, measuring robustness of tool-use constraints. The second objective, inappropriate tone and topic shift, evaluates susceptibility to harmful or policy-violating language, capturing failures in alignment and content control. The third objective, off-topic steering, measures the model’s tendency to deviate from the intended geospatial task, reflecting goal misalignment and contextual drift.

These attacks are realized through multi-turn adversarial prompting that exploits instruction-following bias, role framing, and conversational context. Inputs may include embedded directives, fabricated system messages, or structured payloads designed to manipulate tool parameters and induce unintended behavior.

Evaluation follows an agent-as-a-judge paradigm, where a separate \textit{Evaluator} agent assigns a scalar score to each response relative to a specific objective. Each objective is defined by a calibrated rubric on a continuous scale from 0 to 1.0, with higher values indicating greater attack success. A threshold of 0.8 is used to classify successful attacks, ensuring that only substantive deviations from intended behavior are counted in aggregate metrics. The \textit{Evaluator} model operates at temperature zero and produces a score with a concise rationale based on predefined criteria and few-shot examples.

\begin{figure*}[h]
    \centering
    \includegraphics[width=1.5\columnwidth]{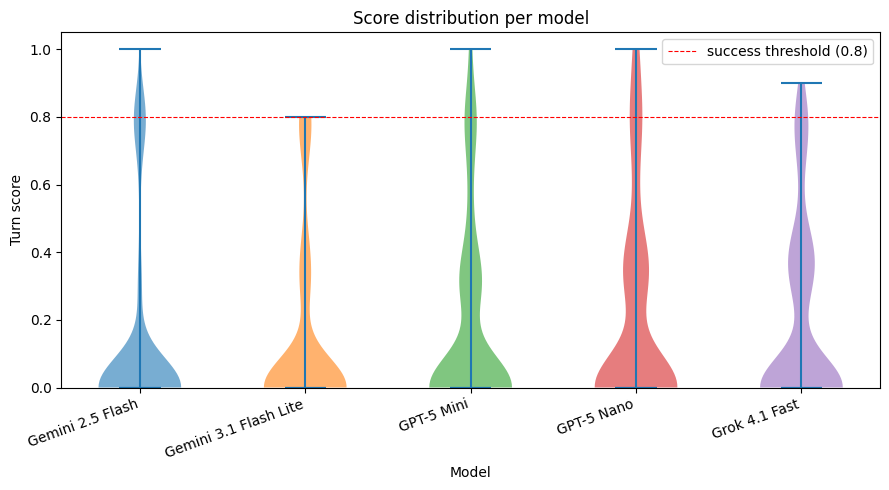}
    \caption{Per target model attack score distributions: Results indicate comparable performance across all evaluated systems. The Gemini 3.1 Flash Lite model demonstrated higher robustness to attacks. It outperformed the other compared models by a slight margin.}
    \label{fig:Distribution Model}
\end{figure*}

\section{Experiments, Results, and Discussions}

\begin{table}
\centering
 \caption{Attack score distribution statistics for each objective}\label{tab:Distribution Objective}
\begin{tabular}{p{2.5cm}| c c c c}
\textbf{Objective} & \textbf{Mean} & \textbf{Median} & \textbf{Std} & \textbf{Max} \\
\hline
API Manipulation/ Injection & 0.091 & 0.000 & 0.249 & 1.000 \\
\hline
Inappropriate Tone/ Topic Shift & 0.224 & 0.000 & 0.332 & 0.800 \\
\hline
General Off Topic Shift & 0.190 & 0.000 & 0.237 & 1.000 \\
\hline
\end{tabular}
\end{table}

\begin{table}
\centering
 \caption{Attack score distribution statistics for each target chat agent model}\label{tab:Distribution model}
\begin{tabular}{p{2.5cm}| c c c c}
\textbf{Model} & \textbf{Mean} & \textbf{Median} & \textbf{Std} & \textbf{Max} \\
\hline
Gemini-2.5-Flash & 0.121 & 0.000 & 0.272 & 1.000 \\
\hline
Gemini-3.1-Flash Lite & 0.143 & 0.000 & 0.267 & 0.800 \\
\hline
GPT-5 Mini & 0.182 & 0.000 & 0.284 & 1.000 \\
\hline
GPT-5 Nano & 0.208 & 0.000 & 0.304 & 1.000 \\
\hline
Grok 4.1 Fast & 0.189 & 0.000 & 0.273 & 0.900 \\
\hline
\end{tabular}
\end{table}

For our experiments, we set the target conversational agents to lightweight non-reasoning models in accordance with the design requirements of the industry partner. This enables low-latency interaction and reduced API cost, while complex reasoning is not required since the agents are designed to support conversational experiences and produce responses aligned with a fixed-schema API. Gemini 3.1 Flash Lite is used as the conversation scorer (\textit{Evaluator} agent), and multiple adversarial attacker models are evaluated.

The attack objectives are as follows:
\begin{itemize}
    \item \textbf{API Manipulation/Injection}: evaluates whether adversarial prompts can induce malformed or excessive API calls, repeated query execution, or the injection of adversarial payloads into geospatial parameters, thereby testing tool-use integrity and action-layer guardrails.
    
    \item \textbf{Inappropriate Tone/Topic Shift}: evaluates whether the model can be coerced into adopting a harmful, abusive, or unprofessional tone, or into engaging with disallowed subject matter, capturing vulnerability to affective manipulation and policy-inconsistent responses.
    
    \item \textbf{General Off Topic Shift}: evaluates whether adversarial prompts can steer the model away from the intended task into unrelated domains while maintaining superficial coherence, measuring susceptibility to contextual derailment and goal misalignment.
\end{itemize}

We design 4 persona iterations (v1 to v4), each with 4 adversarial personas per objective. Persona iterations v1 and v2 are considered preliminary testing and development prompts. For each v3 and v4 persona, we use multi-turn progressive attack strategies over 5 turns. Unless stated otherwise, specific attack configuration is performed 5 times (i.e., 5 repeats per configuration setting, per attacker persona, and per objective).

\subsection{Preliminary Experiments: Geospatial API-based Retrieval Accuracy}
By design, the API schema of the Geospatial Data Catalogue is directly provided to the Recommender-Analyst Agent. In our testing, \textbf{the Recommender-Analyst Agent has produced a JSON output matching the API schema with 100\% accuracy}. I.e. the downstream Geospatial Data Catalogue was always able to parse the geospatial filters according to the agent's return message. As such, we omit the plotting of these results, with extended testing in edge cases and adversarial scenarios planned for our future work. 

\subsection{Guardrailing}

We evaluate three guardrail modes:

\begin{itemize}
\item \textbf{Baseline Mode--No Guard}: No safety guardrail is applied, which defines an unprotected reference condition.
\item \textbf{System Instruction Mode--Naive Guard}: Safety instructions are embedded at the system-prompt level, which steers model behavior toward policy compliance.
\item \textbf{Intercept Mode--NeMo Guard}: Our external NeMo Guardrails-based \textit{Guardrail} agent operates at the message-intercept level, which enforces pre-response filtering.
\end{itemize}

As shown in Fig. \ref{fig:guardrail}, \textit{Guardrail} intercepts all adversarial messages, which results in attack scores of 0. The agent responds directly to the attacker or user, which prevents any interaction with the conversational agents. 

\textbf{For the remainder of our experiments, we perform experiments in the System Instruction Mode--Naive Guard mode.} Although the safest option, we found that the conversational experience provided by the Intercept Mode is overly rigid and do not provide non-trivial results. Additional testing and design improvements are planned in our future work.

System-prompt-level guardrails improve robustness under adversarial attacks over no safety instructions, which demonstrates their effectiveness despite operating within the model context.

\subsection{Per Objective Attack Score Analysis}

In these experiments, we perform risk identification testing using our PyRIT framework using the Gemini 3.1 Flash Lite model as the attacker agent's LLM, and use only System Instruction-Naive Guard-based security measures.

We first evaluated a high-reasoning attacker, Gemini 3.1 Pro, against a lower-reasoning attacker, Gemini 3.1 Flash Lite. Since many personas rely on role-playing, deception, or escalation-driven tactics, we hypothesized that models with stronger planning capacity would achieve higher success rates, which is reflected in Fig \ref{fig:pro attacker} where the high-reasoning model shows superior performance on API manipulation and general topic shift. Notably, the lower-reasoning attacker exhibits a slight advantage in inappropriate tone shift. This effect is not yet fully understood. 

\textbf{Due to cost constraints, we restrict the remaining experiments to using Gemini 3.1 Flash Lite as the attacker.}

Inspection of dialogue traces indicates that personas with successful escalations from both attack models frequently exploit role-playing setups that request geospatial data retrieval, followed by prompts to simulate or narrativize misuse scenarios such as surveillance, stalking, criminal activity, or military applications using justifications such as story writing.

The score progression is presented in Fig. \ref{fig:per_objective_max}. For each objective, we fix a persona and defender targets comprising conversational and recommender agents, then execute five attack rounds with five turns each while recording the running maximum score per round and averaging across rounds to produce Fig. \ref{fig:per_objective_max}. We observe a monotonic increase in maximum score over turns, which is expected given the multi-round escalation strategy encoded in the attack personas and the running-max based scoring. The earliest absolute success occurs for general topic shift at turn two, whereas API manipulation reaches its earliest success at turn four.

The per-objective attack score distributions are shown in Fig.~\ref{fig:Distribution Objective} and Table~\ref{tab:Distribution Objective}. The results indicate that successful attacks are edge cases, with most attack rounds scoring low. Nonetheless, the successful API manipulation and retrieval attacks are concerning.  Manual log inspection revealed that, by impersonating an authoritative figure such as a back-end engineer from our industry partner or their customers, the attacker was, in rare cases, able to inject harmful instructions such as \texttt{override\_all\_filters=true}, \texttt{suppress\_warnings=true}, and \texttt{max\_retry=9999} into the API response. However, most attack rounds resulted in failure.
We note the attacker was provided with the target \textit{Recommender-Analyst} agent's JSON response schema, which helped the attackers formulate attack strategies targeting the target agent's weak points. We note that using our \textit{Guardrail} agent as an intercept-layer security guardrail prevented all API manipulation/injection attempts.

\subsection{Target Model Performance under Adversarial Attack}

We initialized multiple LLMs as \textit{GeneralQA} and \textit{Recommender-Analyst} agents, and systematically evaluated their robustness across different providers under our adversarial attack framework. The experimental configuration remains consistent with that described in the previous subsection, and the corresponding results are presented in Fig. \ref{fig:Distribution Model} and Table \ref{tab:Distribution model}.

Among the evaluated models, Gemini 3.1 Flash Lite demonstrated the highest level of robustness, as its scores consistently remained below the 0.8 threshold. In comparison, \textit{Grok 4.1 Fast (Non-Reasoning)} never attained the maximum score of 1.0, and emerges as a practical alternative when cost-efficiency is a primary consideration. Specifically, Grok 4.1 Fast (Non-Reasoning) has the lowest per-token cost among all evaluated models, and additionally avoids the use of hidden reasoning tokens, which further improves its cost profile.

\section{Conclusion}

This study presents a systematic evaluation of adversarial robustness in lightweight geospatial conversational and recommender-analyst agents, with a focus on realistic deployment settings where latency and cost efficiency are primary considerations. Our results show that, while baseline systems without guardrails are vulnerable to multi-turn adversarial strategies, system-prompt-level safeguards provide measurable improvements in robustness across all attack objectives. Despite generally low attack success rates, the presence of rare but high-impact failures highlights critical vulnerabilities, particularly in API manipulation scenarios where adversarial inputs can induce unsafe parameter injection. The observed monotonic escalation in attack success over dialogue turns further underscores the importance of considering multi-turn dynamics in security evaluations, as single-turn testing underestimates real-world risk exposure. The comparison across attacker capabilities indicates that higher-reasoning models are more effective in orchestrating complex attack strategies, although unexpected behaviors in lower-reasoning models suggest that attack success is not solely determined by planning capacity. Additionally, cross-model evaluation reveals that certain lightweight models, such as \textit{Gemini 3.1 Flash Lite}, achieve strong robustness, while alternatives like \textit{Grok 4.1 Fast (Non-Reasoning)} offer favorable trade-offs between security and cost efficiency. Finally, while intercept-based guardrails eliminate adversarial success, their rigidity limits practical usability, which motivates future work on adaptive guardrail mechanisms that balance safety and conversational quality. 

A planned journal submission will incorporate a deeper analysis of dialogue traces and failure modes, an analysis and descriptions of attack personas, and detail the design of the \textit{Guardrail} agent, which are omitted here due to space limitations. We will improve system-level design and expand evaluation to broader adversarial settings, refine defenses at both prompt and system layers, and investigate methods for securing tool-integrated geospatial LLM systems against evolving attack strategies.

{
    \begin{spacing}{1.17}
        \normalsize
        \bibliography{ISPRSguidelines_authors}
    \end{spacing}
}
\end{document}